\documentclass[10pt,twocolumn,letterpaper]{article}

\usepackage[T1]{fontenc}
\usepackage[pagenumbers]{cvpr}
\usepackage{microtype}
\usepackage{amsmath,amssymb,bm}
\usepackage{graphicx}
\usepackage{booktabs}
\usepackage{xcolor}
\usepackage{enumitem}
\definecolor{cvprblue}{rgb}{0.21,0.49,0.74}
\usepackage[pagebackref,breaklinks,colorlinks,allcolors=cvprblue]{hyperref}

\def\paperID{0000}
\def\confName{CVPR}
\def\confYear{2027}

\graphicspath{{figures/}}
\setlist[itemize]{leftmargin=1.25em,topsep=2pt,itemsep=1pt,parsep=0pt}
\newcommand{\Htwo}{\mathbb H^2}

\newcommand{\Stwo}{\mathbb S^2}
\newcommand{\DR}{D_R}
\newcommand{\method}{\textnormal{\textsc{HyperbolicDiffusion}}}
\newcommand{\hbc}{\textnormal{\textsc{HBC}}}
\newcommand{\meff}{m_{\mathrm{eff}}}

\title{HyperbolicDiffusion: Sharp \& Scalable Tiled Generation on the Hyperbolic Plane}
\author{
Hugo Caselles-Dupr\'e\\
Obvious Research\\
Paris, France
}

\makeatletter
\apptocmd{\@maketitle}{%
  \begin{minipage}{\textwidth}
    \centering
    \includegraphics[width=\linewidth]{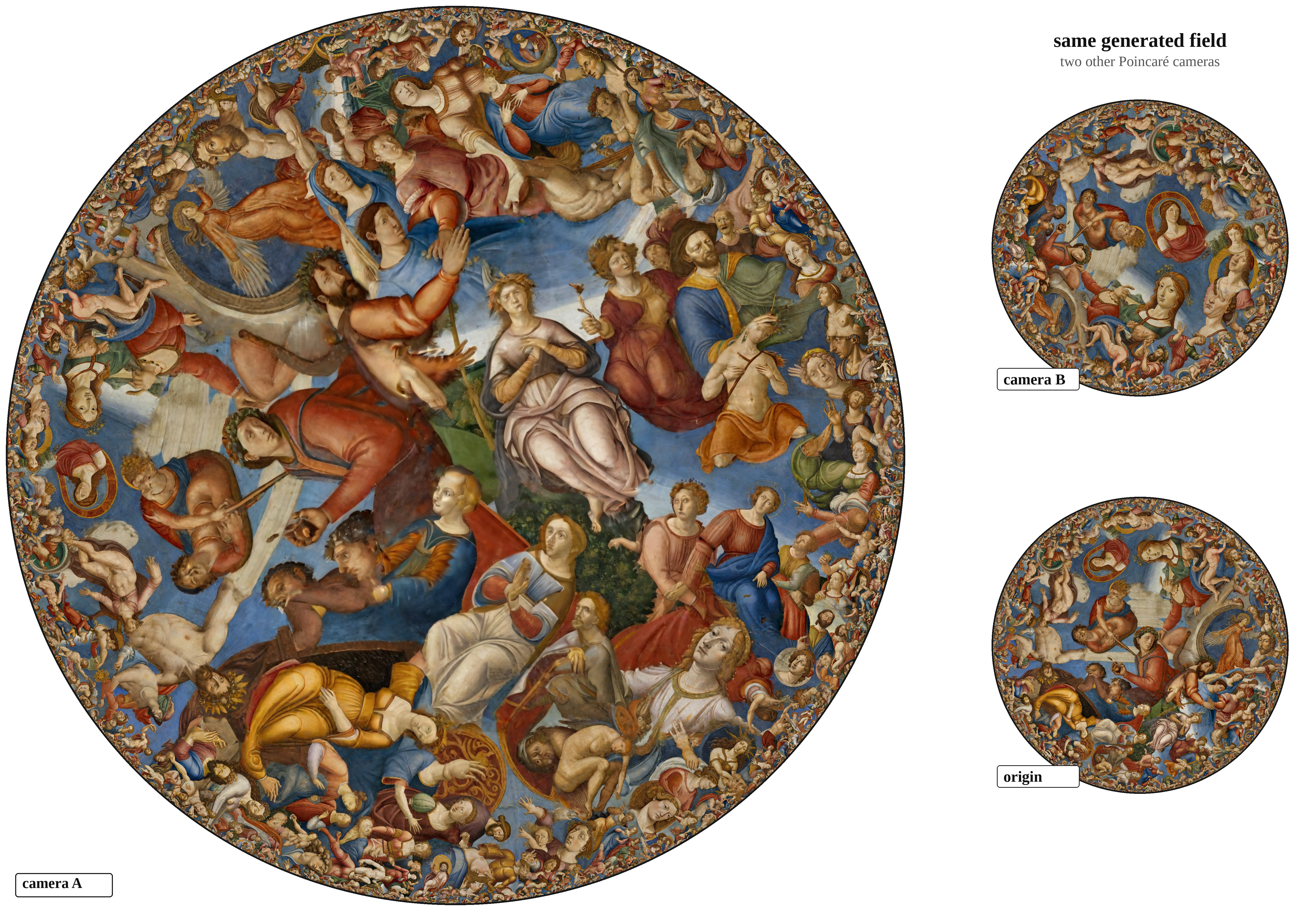}
    \captionof{figure}{\textbf{One generated hyperbolic field, three cameras.}
    The large view and two insets reproject the same 531-window $\Htwo$ field.}
    \label{fig:teaser}
  \end{minipage}%
  \vspace*{5pt}%
}{}{\PackageError{hyperbolicdiffusion}{Could not attach the first-page teaser}{}}
\makeatother

\begin{document}
\maketitle

\begin{abstract}
Planar tiled diffusion denoises overlapping windows of one rectangular
canvas. The hyperbolic plane has no such canvas, and its area grows
exponentially with radius. We introduce \method, a training-free method for
generating finite visual fields directly on the hyperbolic plane $\Htwo$.
Our Hyperbolic Blooming Cover (\hbc{}) reduces window placement to a compact
dynamic program that runs in seconds while providing strong theoretical
guarantees.

\newpage
\vspace*{4pt}
\noindent Permanent surface IDs form a shared latent canvas: a standard
diffusion model denoises local windows, whose predictions are fused back
onto $\Htwo$. Because curvature causes residual disagreement and blur at
multi-window junctions, a geometry-derived second stage re-noises and
repairs precisely those regions. The resulting fields are sharp,
reprojectable, and consistent across viewpoints, providing a prompt-driven
generative counterpart to Escher's \emph{Circle Limit} series.
\end{abstract}

\section{Introduction}

\begin{figure*}[!ht]
\centering
\includegraphics[width=\linewidth]{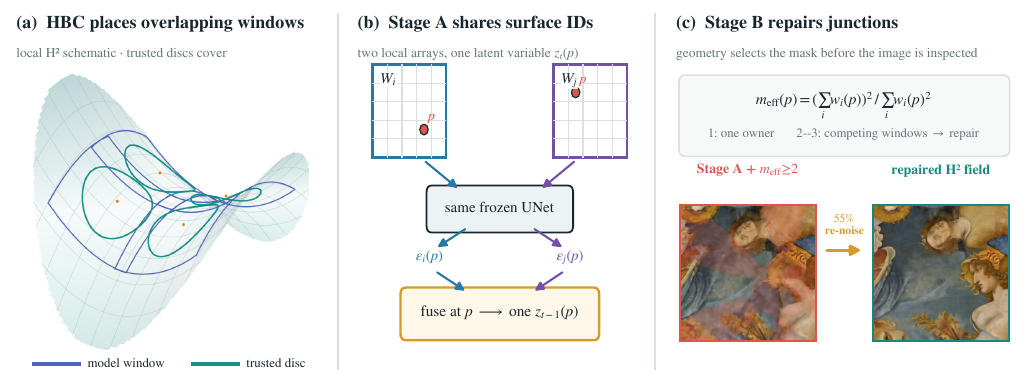}
\caption{\textbf{Method.}
\textbf{(a)} A local negative-curvature schematic shows overlapping model
windows and covering trusted discs; the saddle is illustrative, while the
method uses exponential-map windows on $\Htwo$. \textbf{(b)} Overlapping
arrays share permanent IDs and fuse local denoising proposals in latent
space. \textbf{(c)} Effective multiplicity selects junctions for
$55\%$ re-noise-and-denoise surface repair.}
\label{fig:method}
\end{figure*}

MultiDiffusion generates a large flat image by repeatedly denoising
overlapping crops of one latent array~\cite{bartal2023multidiffusion}. The
overlap works because two crops that contain the same pixel read and update
the same latent variable. We ask whether the same idea can generate an image
that actually lives on the hyperbolic plane, rather than generating a flat
image and warping it into a disc afterward.

This distinction matters. The Poincar\'e disc in Fig.~\ref{fig:teaser} is
only a camera. Once a field has been generated on $\Htwo$, the camera can
move: a motif that was compressed near the rim becomes large at the center,
without regenerating the content. Such a representation can support
navigable hyperbolic murals, non-Euclidean game worlds, VR environments, and
generative analogues of Escher's \emph{Circle Limit} series
~\cite{coxeter1979circle}.

The closest starting point is SphereDiff~\cite{park2026spherediff}. It stores
latent values at points on a sphere, gathers them into local planar views,
and scatters denoising predictions back to the spherical state. However,
sharing a point's latent value does not make the differently warped
two-dimensional neighborhoods seen by each view identical. The views therefore disagree at multi-view junctions, producing
blur or ghosting. Simple texture prompts such as clouds or water
can conceal the defect, while more complex ones, like indoor environments, reveal it.

We adopt SphereDiff's shared latent canvas, and tackle the blur issue. Moreover, $\Htwo$ creates a
new scaling problem that is not present in spherical geometry. The sphere is finite, while a radius-$R$ hyperbolic disc $\DR$
has area
\begin{equation}
 A(\DR)=2\pi(\cosh R-1)=\mathcal{O}(e^R).
 \label{eq:area}
\end{equation}
Every fixed-scale method therefore needs exponentially many windows as $R$
grows. Usually, windows layouts in $\mathbb{R}^2$ or $\mathbb{S}^2$ might be overly wasteful to ensure quality, without sacrificing speed. However, a layout that wastes a modest fraction of views on the plane or sphere can
waste thousands or millions on $\Htwo$. 

\method{} addresses both the exponential placement cost and the junction
quality failure. First,
\hbc{} covers the requested domain with reliable window centers and stores
even a million-center layout as a short list of rings, utilizing a theoretical trick that simplifies the combinatorial problem into a simple dynamic programming one. We provide guarantees, under light assumptions, that the proposed layout by \hbc{} is at most $10\%$ off the optimal solution (which is computationally prohibitive to find). Second, all windows
denoise one field of permanent surface IDs. Third, a geometry-attached
repair stage rewrites the junctions where several differently warped
windows disagree. We empirically demonstrate that this two stages mechanism is the only found mechanism that allows to keep a large composition while removing the blurry junctions. Figure~\ref{fig:method} shows the complete construction.

\section{Related work}

Large-format diffusion methods coordinate local planar generations through
shared variables, factor graphs, or perceptual synchronization
~\cite{bartal2023multidiffusion,zhang2023diffcollage,lee2023syncdiffusion}.
Panorama systems handle periodic boundaries and spherical projections
~\cite{zhang2024panfusion,liu2024panofree}; SphereDiff is closest to our
shared surface representation. Riemannian score models instead define a
generative process for manifold-valued data~\cite{debortoli2022riemannian}.
Here the features are ordinary Euclidean image latents
~\cite{rombach2022ldm} and the denoiser is frozen. Only their
\emph{spatial domain} is curved.

\section{Method}

Let $o$ be the chosen origin and
$\DR=\{x\in\Htwo:d_{\Htwo}(o,x)\leq R\}$ the generated disc. The radius $R$
therefore controls how much of $\Htwo$ is stored. A model invocation uses a
native square mapped with hyperbolic half-width $h$: an edge midpoint is
distance $h$ from its center, while a corner is distance $\sqrt2h$. Smaller
$h$ makes every view locally flatter and usually sharper, but requires more
windows to reach the same $R$. Inside each window, only the central radius
$r_t\leq h$ is trusted for guaranteed coverage. Reducing $r_t/h$ is more
conservative and leaves a larger blending margin, at the cost of more
windows. Thus $R$ sets the requested field, $h$ sets the local model scale,
and $r_t$ sets how much of each window may certify the cover. A model window is a square grid $u\in[-h,h]^2$. At surface center $c_i$,
with local orientation $Q_i$, it reaches the hyperbolic surface through
\begin{equation}
 \chi_i(u)=\exp_{c_i}(Q_i u).
 \label{eq:window}
\end{equation}
In plain terms, $u$ specifies a direction and distance in the flat plane
touching $\Htwo$ at $c_i$; the exponential map follows that direction on the
surface.

\paragraph{Why a simple tiling with overlap fails?} The simplest $\{4,5\}$ tessellation (a tiling of the hyperbolic plane with 5 squares meeting at each junction) covers $\Htwo$, but its cells only touch
edge-to-edge. MultiDiffusion needs overlap, so the relevant baseline keeps
the $\{4,5\}$ centers and enlarges the mapped diffusion windows. Five cells
meet at every tessellation vertex; after enlargement, five differently
warped windows contribute around that same point. These high-multiplicity
junctions are precisely where structured content becomes blurred or
ghosted. Increasing overlap enlarges the five-way junctions, while reducing
the windows enough to suppress them leaves gaps between their reliable
central parts. Thus a $\{4,5\}$ layout or any $\{4,n\}, n\geq5$ does not independently control
reliable coverage and blending multiplicity.

This mapping is not equally faithful throughout the square. At distance
$r$ from its center, tangential lengths are stretched by $\sinh(r)/r$.
The corners, at distance $\sqrt2h$, are therefore the most distorted:
\begin{equation}
\delta(h)=\frac{\sinh(\sqrt2h)}{\sqrt2h}
           =1+\frac{h^2}{3}+O(h^4).
\label{eq:dilation}
\end{equation}
This motivates a conservative coverage unit: the central disc
$B(c_i,r_t)$, with $r_t\leq h$, is the window's \emph{trusted region}.
Every generated point must lie in at least one trusted region. The remainder
of the full square is retained as smooth diffusion overlap, but it is not
used to certify coverage.
\begin{figure*}[!h]
\centering
\includegraphics[width=\linewidth]{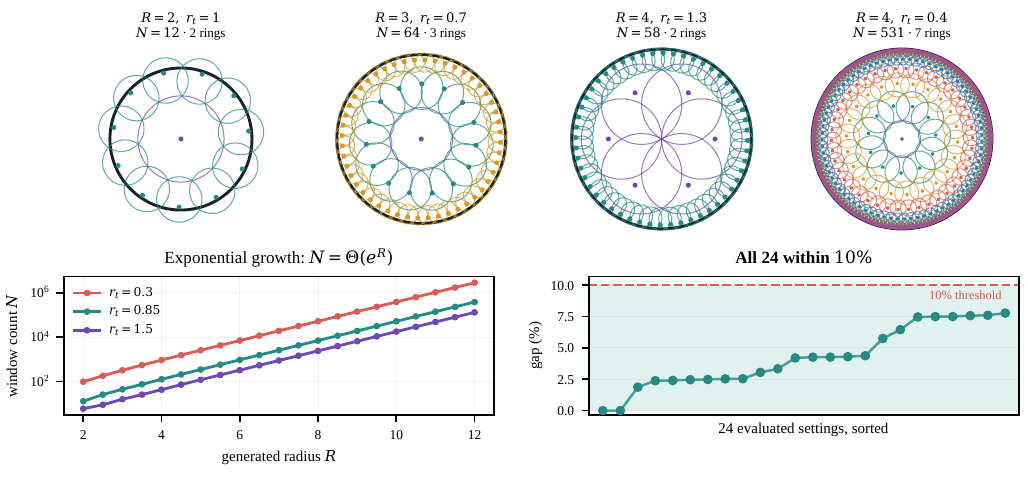}
\caption{\textbf{HBC across scales.}
Four layouts illustrate how HBC adapts across scales. Across 294 tested
settings, every layout has zero holes and the window count follows the
predicted $e^R$ law. In all 24 evaluated cases with $r_t\geq1$, HBC is
within $10\%$ of the optimum.}
\label{fig:generality}
\end{figure*}

\subsection{Hyperbolic Blooming Cover}

Let $C$ be the set of window centers. Following the classical hyperbolic
covering viewpoint~\cite{boroczky2005coverings}, the layout objective is
\begin{equation}
 N^\star(R,r_t)=\min_C |C|
 \quad\text{s.t.}\quad
 \max_{x\in\DR}\min_{c\in C}d_{\Htwo}(x,c)\leq r_t .
 \label{eq:cover}
\end{equation}
Thus $N^\star$ is the smallest number of trusted discs that could cover
$D_R$, without restricting their arrangement. The area ratio already gives
$N^\star=\Omega(e^R)$ for fixed $r_t$.

\hbc{} exploits the radial symmetry of $D_R$. Put $n$ equally spaced
centers on a ring of radius $s$. For points on a target circle of radius
$u$, the worst angular position lies halfway between two centers, and its
distance to the nearest center is exactly
\begin{equation}
 \cosh d_n(u)=\cosh u\cosh s-
 \sinh u\sinh s\cos(\pi/n).
 \label{eq:ring}
\end{equation}
The values of $u$ satisfying $d_n(u)\leq r_t$ form one complete radial
interval. A ring can therefore be represented as an interval with integer
cost $n$. We enumerate feasible rings and solve a one-dimensional weighted
interval cover of $[0,R]$ by dynamic programming. The selected populations
grow outward with the exponentially increasing circumference, so the
trusted discs visually ``bloom'' from the center. The output remains a short
list $(n,s,\phi)$ of ring population, radius, and phase even when expanding
it would produce millions of centers.

The dynamic program is exact within its ring library and every output is
checked for continuous zero-hole coverage. It is not asserted to solve every
finite two-dimensional instance exactly. For fixed $r_t$, however, its
canonical balanced rings obey
\begin{equation}
\limsup_{R\to\infty}\frac{U_{\rm HBC}(R,r_t)}{N^\star(R,r_t)}
\leq
\min\!\left[
1+e^{-2r_t},
\frac{3\sqrt3}{2}\frac{\cosh r_t}{\cosh r_t+1}
\right].
\label{eq:ratio}
\end{equation}
The proven asymptotic error is below $10\%$ for
$r_t\geq\frac12\log10\approx1.151$ and below $5\%$ for
$r_t\geq\frac12\log20\approx1.498$. 

\subsection{Stage A: one shared hyperbolic latent field}

We sample approximately equal-area surface points and assign every point a
permanent latent ID $p$. Each window from the layout cover computed with \hbc{} gathers nearby IDs into the frozen
model's native square array. The same ID can appear in several windows at
different array positions, but it remains one variable.

At every scheduler step, all windows denoise their gathered arrays and
scatter proposals back to the IDs. For each point,
\begin{equation}
 Z_{t-1}(p)=
 \frac{\sum_i w_i(p)\widetilde Z_{i,t-1}(p)}
      {\sum_i w_i(p)+10^{-8}},
\label{eq:fusion}
\end{equation}
which is classic MultiDiffusion, with compact weights implementing the trust region. After the last step, local windows gather
the final IDs and are VAE-decoded. The Poincar\'e projection never enters
the denoising state; it is only a way to inspect geometry and render the
stored field.

\subsection{Stage B: Geometry-attached junction repair}

Shared IDs solve only half of the correspondence problem. If three windows
contain the same point, they agree on its central latent value. They do not
present the denoiser or VAE with the same two-dimensional neighborhood:
curvature and window orientation arrange the nearby IDs differently.
So, the windows can
decode slightly different faces, edges, or ornaments. Their mixture becomes
blur or ghosting where several windows contribute strongly.

We locate these regions from geometry alone. If $\bar w_i(p)$ is window
$i$'s normalized weight at surface point $p$, define
\begin{equation}
 \meff(p)=\frac{1}{\sum_i\bar w_i(p)^2}.
 \label{eq:meff}
\end{equation}
$\meff\approx1$ means one window effectively owns the point; values near two
or three mean that several windows contribute comparably. Empirically, the mask created by $\meff\geq2$ consistently aligns with the visible blur.

We thus introduce Stage B which repairs the field, not a particular screenshot. It covers
the mask with larger surface windows, reconstructs and VAE-encodes a coherent
Stage-A image in each, then adds scheduler noise at strength $0.55$ and
denoises only the mask with the same prompt. Then, MultiDiffusion operates
inside each repair window; one final owner writes every repaired point back
to $\Htwo$. Because each window stores its center, orientation, and ownership
region, the correction moves when the camera is recentered. It is not
screen-space inpainting.

\paragraph{Why the two stages are separate?} The crucial aspect of using two stages in that re-encoding after Stage A allows to control the density of the shared hyperbolic latent field. Interpolation on the field is not possible due to loss of Gaussian properties of the latent field. Five one-stage families failed to match the sequential result: denser bases;
late insertion of fine IDs; joint coarse/fine states from initial noise;
hard or blended ownership; and synchronized or wavelet decoding.

\begin{figure*}[!ht]
\centering
\includegraphics[width=\linewidth]{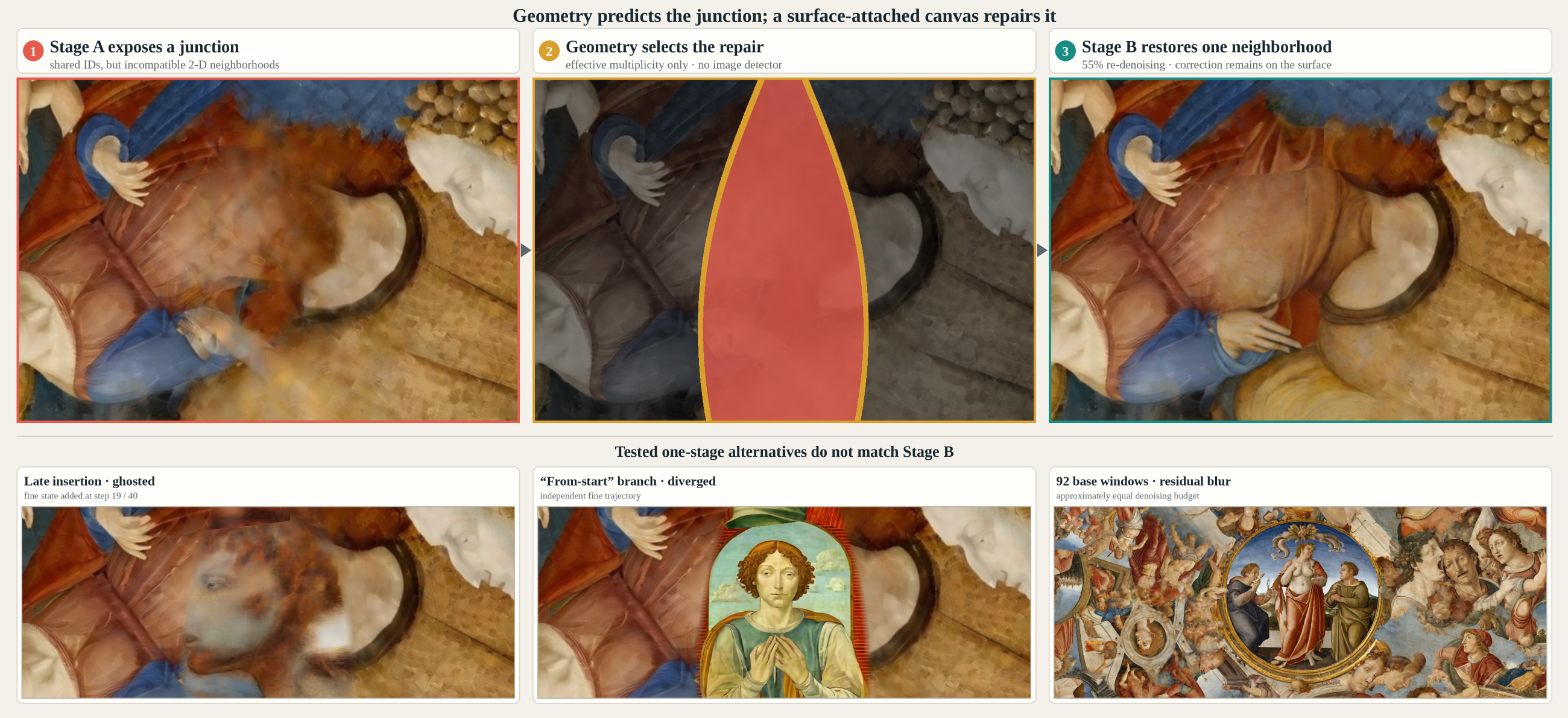}
\caption{\textbf{Why a second stage is needed.}
The geometry-only multiplicity mask follows Stage-A junction blur. Re-noising
and denoising these surface regions at 55\% restores coherent detail.
Matched one-stage controls remain ghosted, diverge, or retain residual blur.}
\label{fig:repair}
\end{figure*}

\section{Evaluation}

\paragraph{Broad behavior of HBC.}
We test 294 settings with $R\in[2,12]$ and $r_t\in[0.15,2]$. Every HBC
layout provides complete coverage, and the number of windows grows as
$e^R$, as predicted by our analysis. Planning takes only $0.45$ seconds
on average. Even the largest layout, containing 11.4 million windows,
is computed in 5.44 seconds.

We evaluate HBC on 24 settings with $r_t \geq 1$. In every case, its
window count is provably within $10\%$ of the optimum. Together with Eq.~\eqref{eq:ratio}, this demonstrates that HBC is a fast
and consistently near-optimal placement method across the evaluated
regime.

\paragraph{Image quality and repair.}
Tests span frescoes, architecture, natural textures, several seeds,
$h\in[0.471,1]$, radii up to $4$, and layouts from tens to 531 windows.
Smaller $h$ improves Stage A, but structured content still exposes
high-multiplicity junctions. In our experiments, Stage B raises masked
mean gradient magnitude by an average of $15\%$ and masked Laplacian
magnitude by an average of $17\%$. Because ghost edges can also raise these
metrics, qualitative evaluation remains the most important metric. The aligned crops in Fig.~\ref{fig:repair} show how efficient Stage B is.

\paragraph{SphereDiff stress test.}
We reproduce SphereDiff and test it on harder prompts than in their paper. Texture prompts are forgiving:
clouds, water, and light contain few rigid structures whose continuity
reveals a junction. A normal indoor scene is a harder test and exposes
washed bands, blurry zones, as predicted.
Our Stage B transfers to $\Stwo$ by replacing hyperbolic distances and
windows with their spherical counterparts. It qualitatively outperforms the released Stage-A
construction on this structured stress test.

\paragraph{Reprojection and interactive navigation.}
Figure~\ref{fig:teaser} renders one saved field from three centers without
repeating generation or repair. Our WebGL prototype streams decoded windows
and evaluates the Poincar\'e projection as the camera moves; it runs
interactively in desktop-browser tests. This supports navigable murals,
game worlds, and VR textures, although we do not report headset latency or a
user study.

\section{Conclusion}

\method{} combines HBC, shared-ID latent fusion, and repair
for sharp, movable generation on $\Htwo$. Future work will tighten finite
certificates and seek a single multiresolution diffusion state.

{\small
\bibliographystyle{ieeenat_fullname}
\bibliography{references}

@inproceedings{bartal2023multidiffusion,
  title     = {{MultiDiffusion}: Fusing Diffusion Paths for Controlled Image Generation},
  author    = {Bar-Tal, Omer and al.},
  booktitle = {Proceedings of the 40th International Conference on Machine Learning},
  series    = {Proceedings of Machine Learning Research},
  volume    = {202},
  pages     = {1737--1752},
  publisher = {PMLR},
  year      = {2023}
}

@inproceedings{park2026spherediff,
  title     = {{SphereDiff}: Tuning-Free 360-Degree Static and Dynamic Panorama Generation via Spherical Latent Representation},
  author    = {Park, Minho and Kang, Taewoong and Yun, Jooyeol and Hwang, Sungwon and Choo, Jaegul},
  booktitle = {Proceedings of the AAAI Conference on Artificial Intelligence},
  volume    = {40},
  number    = {10},
  pages     = {8305--8313},
  year      = {2026},
  doi       = {10.1609/aaai.v40i10.37779}
}

@inproceedings{zhang2023diffcollage,
  title     = {{DiffCollage}: Parallel Generation of Large Content with Diffusion Models},
  author    = {Zhang, Qinsheng and al.},
  booktitle = {Proceedings of the IEEE/CVF Conference on Computer Vision and Pattern Recognition},
  pages     = {10188--10198},
  year      = {2023},
  doi       = {10.1109/CVPR52729.2023.00982}
}

@inproceedings{lee2023syncdiffusion,
  title     = {{SyncDiffusion}: Coherent Montage via Synchronized Joint Diffusions},
  author    = {Lee, Yuseung and Kim, Kunho and Kim, Hyunjin and Sung, Minhyuk},
  booktitle = {Advances in Neural Information Processing Systems},
  volume    = {36},
  pages     = {50648--50660},
  year      = {2023},
  doi       = {10.52202/075280-2203}
}

@inproceedings{zhang2024panfusion,
  title     = {Taming Stable Diffusion for Text to 360 Panorama Image Generation},
  author    = {Zhang, Cheng and al.},
  booktitle = {Proceedings of the IEEE/CVF Conference on Computer Vision and Pattern Recognition},
  pages     = {6347--6357},
  year      = {2024}
}

@inproceedings{liu2024panofree,
  title     = {{PanoFree}: Tuning-Free Holistic Multi-View Image Generation with Cross-View Self-Guidance},
  author    = {Liu, Aoming and Li, Zhong and Chen, Zhang and Li, Nannan and Xu, Yi and Plummer, Bryan A.},
  booktitle = {Computer Vision -- ECCV 2024},
  series    = {Lecture Notes in Computer Science},
  volume    = {15085},
  pages     = {146--164},
  publisher = {Springer},
  year      = {2024},
  doi       = {10.1007/978-3-031-73383-3_9}
}

@inproceedings{debortoli2022riemannian,
  title     = {Riemannian Score-Based Generative Modelling},
  author    = {De Bortoli, Valentin and al.},
  booktitle = {Advances in Neural Information Processing Systems},
  volume    = {35},
  pages     = {2406--2422},
  year      = {2022},
  doi       = {10.52202/068431-0175}
}

@article{boroczky2005coverings,
  title   = {Finite Coverings in the Hyperbolic Plane},
  author  = {B{\"o}r{\"o}czky, Jr., K{\'a}roly},
  journal = {Discrete \& Computational Geometry},
  volume  = {33},
  pages   = {165--180},
  year    = {2005},
  doi     = {10.1007/s00454-004-1101-y}
}

@article{coxeter1979circle,
  title   = {The Non-Euclidean Symmetry of Escher's Picture ``Circle Limit III''},
  author  = {Coxeter, H. S. M.},
  journal = {Leonardo},
  volume  = {12},
  number  = {1},
  pages   = {19--25},
  year    = {1979},
  doi     = {10.2307/1574078}
}

@inproceedings{rombach2022ldm,
  title     = {High-Resolution Image Synthesis with Latent Diffusion Models},
  author    = {Rombach, Robin and Blattmann, Andreas and Lorenz, Dominik and Esser, Patrick and Ommer, Bj{\"o}rn},
  booktitle = {Proceedings of the IEEE/CVF Conference on Computer Vision and Pattern Recognition},
  pages     = {10684--10695},
  year      = {2022},
  doi       = {10.1109/CVPR52688.2022.01042}
}
}

\end{document}